\documentclass[12pt]{article}    
\usepackage{authblk}
\usepackage{fullpage} 

\usepackage{natbib}
\bibpunct{(}{)}{;}{a}{,}{,}

\usepackage{amsmath}
\usepackage{enumitem}
\usepackage{natbib}
\usepackage{color}
\usepackage{booktabs}

\newtheorem{theorem}{Theorem}
\newtheorem{lemma}{Lemma}
\newtheorem{assumption}{Assumption}
\newtheorem{proof}{Proof}

\usepackage{algorithm}
\usepackage{algorithmic}
\usepackage{multirow}
\usepackage{amsmath, amssymb}
\usepackage{setspace}
\usepackage{caption}

\usepackage{graphicx}
\graphicspath{ {images/} }
\usepackage{float}

\usepackage{hyperref}

\newcommand{\Ex}[1]{\mathbb{E}(#1)}

\newcommand{\ISGDhalf}{{\tt ISGD}$^{1/2}$}
\newcommand{\SGDhalf}{{\tt SGD}$^{1/2}$}
\newcommand{\Le}{E}

\newcommand{\comment}[1]{\text{ }\hspace{20px} \text{[\small { #1 }]}}
\newcommand\numberthis{\addtocounter{equation}{1}\tag{\theequation}} 
\newcommand{\paperTitle}{Convergence diagnostics for stochastic gradient descent
with constant learning rate}

\newcommand{\SGDTheorem}{
\begin{theorem}[\citep{moulines2011non, needell2014stochastic}]
\label{theorem:sgd}
Under certain assumptions on the loss function, there are positive constants 
$A_\gamma, B$ such that, for every $n$,  it holds that
$$
\Ex{||\theta_n - \theta_\star||^2} \le \Ex{||\theta_0-\theta_\star||^2} e^{-A_\gamma n}  + B \gamma.
$$
\end{theorem}
}

\newcommand{\TheoremPflugFinite}{
\begin{theorem}
\label{theorem:pflugFinite}
Consider SGD with constant rate,
$$
\theta_n = \theta_{n-1} - \gamma \nabla \ell(y_n, x_n^\top \theta_{n-1}).
$$
Suppose that Theorem~\ref{theorem:sgd} holds, so that that $\Ex{||\theta_n - \theta_\star||^2} \le \gamma M$, 
for some positive $M$ and large enough $n$.
We make the following additional assumptions:
\begin{enumerate}
\item[(a)] $\nabla \ell(y, x^\top \theta) = f(x, \theta) + e$, where $f(x, \theta)$ is $L$-Lipschitz, $\Ex{e | x, \theta} = 0$ 
and $\Ex{||e||^2} \ge \tau^2$.
\item[(b)] It holds $\Ex{f(x, \theta -\gamma z}^\top  z) \le \Ex{f(x, \theta)^\top  z} - \gamma K \cdot \Ex{z^\top  C z}$, 
for any $\theta, z$,
for some positive constant $K$, and some positive definite matrix $C$ with minimum eigenvalue $\mu > 0$.
\item[(c)] 
It holds that
 $\gamma > (L^2 M - \mu K \tau^2) /\mu K L^2 M$.
\end{enumerate}
Then, 
$$
\Ex{\nabla \ell(y_{n}, x_{n}^\top \theta_{n-1}}^\top  \nabla \ell(y_{n+1}, x_{n+1}^\top \theta_n)) < 0.
$$
\end{theorem}
}

\newcommand{\ProofPflugFinite}{
\begin{proof}
For brevity let $\tilde{\ell}_i = f(x_{i+1}, \theta_{i}) + e_i = f_i + e_i$ be the stochastic gradient at iteration $i+1$.

\begin{align}
\Ex{\tilde{\ell}_{i-1}^\top \tilde{\ell}_i} 
& = \mathbb{E}\left[(f_{i-1} + e_{i-1})^\top  (f_i + e_i)\right]= \mathbb{E}\left[(f_{i-1} + e_{i-1})^\top  f_i \right]
\comment{because $e_i$ are zero-mean}\nonumber\\
& = \mathbb{E}\left[ (f_{i-1} + e_{i-1})^\top  f\left(\theta_{i-1} - \gamma f_{i-1} -\gamma e_{i-1}\right)\right]
\comment{by  SGD step for $
\theta_i$}\nonumber\\
& \le \Ex{||f_{i-1}||^2}  - \gamma K \cdot \mathbb{E}\left[(f_{i-1} + e_{i-1})^\top  C (f_{i-1} + e_{i-1})\right] 
\comment{by Assumption~(b)}\nonumber\\
& \le (1-\gamma \mu K) \Ex{||f_{i-1})||^2} - \gamma K \cdot \Ex{||e_{i-1}||_C^2}
\nonumber\\
&  \le (1-\gamma \mu K) L^2 \Ex{||\theta_{i-1}-\theta_\star||^2} - 
\gamma \mu K \tau^2
\comment{ by Lipschitz Assumption~(a)}\nonumber\\
& \le \gamma[(1-\gamma \mu K) L^2 M - \mu K \tau^2]
\nonumber\\
& < 0.
\comment{by Assumption~(c) and small enough $\gamma$} 
\end{align}
\end{proof}
}

\newcommand{\LinearTheorem}{
\begin{theorem}
\label{theorem:linear}
Suppose that the loss is quadratic, $\ell(y, x^\top\theta) = (1/2) (y-x^\top \theta)^2$. Let $x_1$ and $x_2$ be two iid vectors from the distribution of $x$, and define: 
$\sigma^2 =\Ex{(y-x^\top\theta_\star)^2} $; 
$c^2 = \Ex{(x_1^\top x_2)^2} $; 
$C = \Ex{x_1x_2^\top (x_1^\top x_2)}$;
$D = \Ex{x_1 x_1^\top (x_1^\top x_2)^2}$, and suppose that all such constants are finite. Then, for $\gamma >0$,
\begin{align}
\Delta_n(\theta) & = \Ex{S_{n+2} -  S_{n+1} | \theta_{n} = \theta} \nonumber\\
& = (\theta-\theta_\star)^\top (C-\gamma D) (\theta-\theta_\star)- \gamma c^2 \sigma^2.\nonumber
\end{align}
\end{theorem}
}
\newcommand{\LinearProof}{
\begin{proof}
\label{proof:linear}
For notational brevity we make the following definitions:
\begin{align}
\label{eq:two_steps}
\theta^{+} 
& = \theta + \gamma (y_1 - x_1^\top\theta) x_1\nonumber\\
\theta^{++} & = \theta^+ + \gamma (y_2 - x_2^\top\theta^+) x_2,
\end{align}
where $\theta$ is the current iterate, and $\theta^+$ and $\theta^{++}$ are the next two using iid data $(x_1, y_1)$ and $(x_2, y_2)$. For a fixed $\theta$ 
we understand the Pflug diagnostic through the function
\begin{align}
H(\theta) &= S_{++} - S_{+} | \theta = \nabla_{++} \ell^\top \nabla_{+} \ell = (\theta^+ - \theta)^\top(\theta^{++} - \theta^+) / \gamma^2 \\ 
\textit{ and } \Delta_{n}(\theta) &= \Ex{H(\theta)} = \mathbb{E}\left((\theta^+ - \theta)^\top(\theta^{++} - \theta^+) / \gamma^2\right).
\end{align}

We use Equation~\eqref{eq:two_steps} to derive an expression for $H$:
\begin{align}
\label{eq:H_ex}
H(\theta) & = (y_1 - x_1^\top\theta) (y_2 - x_2^\top\theta^+) x_1^\top x_2\nonumber\\
& = (y_1 - x_1^\top\theta) \left[y_2 - x_2^\top\theta - \gamma (y_1 - x_1^\top\theta) x_1^\top x_2\right] x_1^\top x_2\nonumber\\
& = (y_1 - x_1^\top\theta) (y_2 - x_2^\top\theta) x_1^\top x_2  - 
\gamma (y_1 - x_1^\top\theta)^2 (x_1^\top x_2)^2.
\end{align}

Let $y_i = x_i^\top\theta_\star + \varepsilon_i$; we know that $\Ex{(y_i - x_i^\top\theta_\star) x_i} = 0$.
Now, we analyze each term individually:
\begin{align}
(y_1 - x_1^\top\theta) (y_2 - x_2 ^\top\theta) x_1^\top x_2
 & = [x_1^\top (\theta_\star-\theta) + \varepsilon_1] [x_2^\top (\theta_\star-\theta) + \varepsilon_2] x_1^\top x_2\nonumber\\
 & = (\theta - \theta_\star)^\intercal x_1 x_2^\top (x_1^\top x_2) (\theta-\theta_\star)
  + \varepsilon_1 W^{(1)} + \varepsilon_2 W^{(2)} + \varepsilon_1\varepsilon_2 W^{(3)}.\nonumber
\end{align}
The $W$ variables are conditionally independent of $\varepsilon$ and so using the law of iterated expectations these terms vanish. 
$$
\mathbb{E}\left((y_1 - x_1^\top\theta) (y_2 - x_2 ^\top \theta) x_1^\top x_2\right)
=  (\theta - \theta_\star)^\intercal \mathbb{E}\left(x_1 x_2^\top  (x_1^\top x_2)\right) (\theta-\theta_\star)
= (\theta - \theta_\star)^\intercal C (\theta-\theta_\star).$$

Using a similar reasoning, for the second term we have:
\begin{align}
\label{eq:second}
(y_1 - x_1^\top \theta)^2 (x_1^\top x_2)^2
 = & \left[ (x_1^\top (\theta_\star - \theta) + \varepsilon_1  \right]^2 (x_1^\top x_2)^2\nonumber\\
 = &  (\theta - \theta_\star)^\intercal  x_1 x_1^\top  (x_1^\top x_2)^2 (\theta-\theta_\star)
 + \varepsilon_1 W^{(4)} + \varepsilon_1^2 (x_1^\top x_2)^2.
\end{align}
In expectation of Equation~\eqref{eq:second},
\begin{align}
\label{eq:third}
\mathbb{E}\left((y_1 - x_1^\top \theta)^2 (x_1^\top x_2)^2\right)
 & =  (\theta - \theta_\star)^\intercal \Ex{x_1x_1^\top  (x_1^\top x_2)^2} (\theta-\theta_\star) + \varepsilon_1^2 (x_1^\top x_2)^2\nonumber\\
 & = (\theta - \theta_\star)^\intercal D (\theta-\theta_\star) 
 + \sigma^2 c^2.
 \end{align}
 
By combining all results we finally get:
$$
\Delta_{n}(\theta) =(\theta - \theta_\star)^\intercal  (C - \gamma D) (\theta-\theta_\star)
- \gamma\sigma^2c^2.$$

\end{proof}
}

\newcommand{\LinearTheoremImplicit}{
\begin{theorem}
\label{theorem:linear_implicit}
Let $\lambda_\gamma = \Ex{1 / (1 + \gamma ||x||^2)} \in (0, 1]$.
Under the assumptions of Theorem~\ref{theorem:linear} applied on the implicit procedure in Equation~\eqref{eq:implicit}, it holds that
\begin{align}
\Delta_n^{\mathrm{im}}(\theta) & = \Ex{S_{n+2} - S_{n+1} | \theta_{n} = \theta} \nonumber\\
& = a_\gamma \Delta_n(\theta) + b_\gamma\left[(\theta-\theta_\star)^\top D (\theta-\theta_\star)
+ \sigma^2 c^2\right],\nonumber
\end{align}
where $a_\gamma = \lambda_\gamma^2$, 
$b_\gamma=\gamma \lambda_\gamma^2 (1-\lambda_\gamma)$.\end{theorem}
}
\newcommand{\LinearProofImplicit}{
\begin{proof}
\label{proof:linear_implicit}

We derive similar theoretical results for $H^{im} (\theta), \Delta_{n}^{im}(\theta)$ under the linear normal model for implicit updates. We have the implicit updates
\begin{align*}
\theta^{+}   &= \theta + \gamma ( y_{1} - x_{1}^\top  \theta^{+} ) x_{1} \\
\theta^{++} &= \theta^{+} + \gamma ( y_{2} - x_{2}^\top  \theta^{++} ) x_{2}
\end{align*}
Also note the collinearity 
\begin{align*}
( y_{1} - x_{1}^\top  \theta^{+} )    &= \lambda_{1} ( y_{1} - x_{1}^\top  \theta )  \\
( y_{2} - x_{2}^\top  \theta^{++} ) &= \lambda_{2} ( y_{2} - x_{2}^\top  \theta^{+} ),\\\nonumber
& = \lambda_2 [y_2 - x_2^\top \theta - \gamma \lambda_1(y_1-x_1^\top \theta) x_1^\top x_2],
\end{align*}
where $\lambda_1 = 1 / (1+ \gamma ||x_1||^2)$ and 
$\lambda_2 = 1 / (1+ \gamma ||x_2||^2)$.
We derive an expression for $H^{im}$, with implicit updates:
\begin{align*}
H^{im} (\theta) &= (\theta^+ - \theta)^\top (\theta^{++} - \theta^+) / \gamma^2 \\
&= ( y_{1} - x_{1}^\top  \theta^{+} ) ( y_{2} - x_{2}^\top  \theta^{++} ) x_{1}^\top  x_{2} \\
&= \lambda_{1} \lambda_{2} ( y_{1} - x_{1}^\top  \theta ) [y_2 - x_2^\top \theta - \gamma \lambda_1(y_1-x_1^\top \theta) x_1^\top x_2] x_1^\top  x_2 \\
& =  \lambda_1\lambda_2\left[ H(\theta)
+ \gamma(1-\lambda_1) (y_1-x_1^\top \theta)^2 (x_1^\top x_2)^2\right] \\
&= \lambda_1\lambda_2 H(\theta) + \gamma\lambda_1\lambda_2 (1-\lambda_1) (y_1-x_1^\top \theta) (x_1^\top x_2)^2,
\end{align*}
where $H$ is the function from the explicit update in Equation~\eqref{eq:H_ex}. The formula for $\Delta_n^{im}(\theta)$ follows by applying expectation and the reasoning in Equation~\eqref{eq:third}. Note that $\Ex{\lambda_1\lambda_2} = 
\lambda_\gamma^2$ since $\lambda_1$ and $\lambda_2$ are independent and have marginally identical distributions.
\end{proof}
}

\newcommand{\TheoremGeneral}{
\begin{theorem}
\label{theorem:general}
Define the loss $\ell(y, x^\top\theta) = -y\cdot x^\top\theta + f(x^\top\theta)$. Let $h(u) = f'(u)$ and suppose that $h'(x^\top\theta) \ge k > 0$, almost surely for all $\theta$.
Let $x_1, x_2$ be two iid vectors from the distribution of $x$. 
Define $\sigma^2 = \Ex{(y-h(x^\top\theta_\star)^2}$; 
$c^2 = \Ex{(x_1^\top x_2)^2}$;
$C(\theta, \theta_\star) = 
\Ex{ [h(x_1^\top\theta) - h(x_1^\top\theta_\star)] x_1}$;
$D^2(\theta, \theta_\star)
= \Ex{[h(x_1^\top\theta) - h(x_1^\top\theta_\star)]^2 (x_1^\top x_2)^2}$.
Then, for small enough $\gamma$,
\begin{align}
\Delta_n^{glm}(\theta) & = \Ex{S_{n+2} - S_{n+1} | \theta_{n} = \theta} \nonumber
\le 
||C(\theta, \theta_\star)||^2 - \gamma k [\sigma^2 c^2 + D^2(\theta, \theta_\star)].
\nonumber
\end{align}
\end{theorem}
}
\newcommand{\ProofGeneral}{
\begin{proof}
\label{proof:general}
The updates for the GLM loss are as follows:
\begin{align}
\label{eq:two_steps_h}
\theta^{+} 
& = \theta + \gamma (y_1 - h(x_1^\top \theta)) x_1\nonumber\\
\theta^{++} & = \theta^+ + \gamma (y_2 - h(x_2^\top  \theta^+)) x_2,
\end{align}
Note that $h(x_2^\top\theta^+) = 
h(x_2^\top\theta) + \gamma h'(x_2^\top\theta) (y_1 - h(x_1^\top\theta))x_2^\top x_1 + O(\gamma^2)$. 
We can now follow the exact same reasoning as in Theorem~\ref{theorem:linear} and that $h'(x^\top\theta) \ge k$ almost 
surely.

\end{proof}
}

\newcommand{\Assumptions}{
\begin{assumption}
The following assumptions are true with regard to procedure in Equation~\eqref{eq:implicit}.
\begin{enumerate}[label = (\alph*)]
\item Function $\ell$ is convex, twice differentiable almost surely with respect to $x^\top \theta$. 
\item For the observed Fisher information matrix $\hat{ \mathcal{I} }_{n} (\theta) = \nabla^{2} \ell (y_n, x_n^\top \theta)$ there exists constants $b > 0$ and $0 < t < \infty$ such that $b \leq trace( \hat{ \mathcal{I} }_{n} (\theta) ) \leq t$ almost surely, for all $\theta$. 
The Fisher information matrix $\mathcal{I} (\theta_{*}) = \mathbb{E} \left( \hat{ \mathcal{I} }_{n} (\theta_{*}) \right)$ has minimum eigenvalue $\lambda > 0$.
\item There exists $\sigma^{2} >0$ such that, for all $n$, $\mathbb{E} ( \| \nabla \ell (y_n, x_n^\top \theta_\star \|^{2} | \mathcal{F}_{n-1} ) \leq \sigma^{2}$, almost surely.
\item The function $\theta \mapsto \mathbb{E} ( \nabla \ell(y,  x^\top 
\theta) )$ is Lipschitz with constant $L$, i.e., for all $n, \theta_1, \theta_2$,
\begin{align*}
&\mathbb{E} ( \| \nabla \ell(y_{n}; x_{n}^\top \theta_{1}) - \nabla \ell(y_{n}; x_{n}^\top  \theta_{2}) \|^{2} | \mathcal{F}_{n-1} ) 
\leq L^{2} \| \theta_{1} - \theta_{2} \|^{2}.
\end{align*}
\item Learning rate $\gamma>0$ is such that
$
\gamma L^{2} (1 + \gamma t) < \lambda (1 + \gamma b)^2.
$
\end{enumerate}
\end{assumption}
}

\newcommand{\TheoremGradient}{
\begin{lemma}
\label{lemma:gradient}
The gradient $\nabla \ell$ is a scaled version of covariate $x$, i.e., for every $\theta \in \mathbb{R}^{p}$ there is a scalar $\lambda \in \mathbb{R}$ such that
\begin{align*}
\nabla \ell (y; x^\top  \theta) = \lambda x
\end{align*}

Thus, the gradient in the implicit update is a scaled version of the gradient calculated at the previous iterate, i.e.,
\begin{align*}
\nabla \ell (y_{n}; x_{n}^\top  \theta_{n}) = \lambda_{n} \nabla \ell (y_{n}; x_{n}^\top  \theta_{n-1}), 
\numberthis \label{scaled-likelihood}
\end{align*}

where the scalar $\lambda_{n}$ satisfies
\begin{align*}
\lambda_{n} \ell' (y_{n}; x_{n}^\top  \theta_{n-1}) = 
\ell' (y_{n}; x_{n}^\top  \theta_{n-1} + \gamma \lambda_{n} \ell' (y_{n}; x_{n}^\top  \theta_{n-1}) x_{n}^\top  x_{n})
\numberthis \label{scaled-fixedpoint}
\end{align*}
\end{lemma}
}

\newcommand{\ProofGradient}{
\begin{proof}
From the chain rule 
$
\nabla \ell (y_{n}; x_{n}^\top  \theta_{n}) = \ell' (y_{n}; x_{n}^\top  \theta_{n}) x_{n}
$
, and similarly 
$
\nabla \ell (y_{n}; x_{n}^\top  \theta_{n-1}) = \ell' (y_{n}; x_{n}^\top  \theta_{n-1}) x_{n}
$.
Thus the two gradients are collinear. Therefore there exists a scalar $\lambda_{n}$ such that
\begin{align*}
\ell' (y_{n}; x_{n}^\top  \theta_{n}) x_{n} 
&= \lambda_{n} \ell' (y_{n}; x_{n}^\top  \theta_{n-1}) x_{n}
\numberthis \label{scaled-proof-1}
\end{align*}

We also have,
\begin{align*}
\theta_{n} &= \theta_{n-1} + \gamma \nabla \ell (y_{n}; x_{n}^\top  \theta_{n})
\ [ \textit{by definition of implicit SGD update Equation~}\eqref{eq:implicit}] \\
&= \theta_{n-1} + \gamma \lambda_{n} \ell ' (y_{n}; x_{n}^\top  \theta_{n-1}) x_{n}
\ [ \textit{by chain rule and Equation} \eqref{scaled-proof-1}]
\numberthis \label{scaled-proof-2}
\end{align*}

Substituting the expression for $\theta_{n}$ in Equation\eqref{scaled-proof-2} into Equation\eqref{scaled-proof-1} we obtain the desired result of the theorem. From Equation\eqref{scaled-proof-1} we get the equality
\begin{align*}
\ell' (y_{n}; x_{n}^\top  \theta_{n}) = \lambda_{n} \ell' (y_{n}; x_{n}^\top  \theta_{n-1})
\numberthis \label{scaled-proof-3}
\end{align*}

and substituting we get our desired result
\begin{align*}
\lambda_{n} \ell' (y_{n}; x_{n}^\top  \theta_{n-1}) 
&= \ell ' (y_{n} ; x_{n}^\top  ( \theta_{n-1} + \gamma \lambda_{n} \ell ' (y_{n}; x_{n}^\top  \theta_{n-1}) x_{n} ) ) \\
&= \ell ' (y_{n}; x_{n}^\top  \theta_{n-1} + \gamma \lambda_{n} \ell ' (y_{n}; x_{n}^\top  \theta_{n-1}) x_{n} ^\top  x_{n})
\end{align*}
\end{proof}
}

\newcommand{\LemmaLambda}{
\begin{lemma}
\label{lemma:gradBd}
Suppose Assumptions 1 (a), and (b) hold. Then, almost surely it holds
\begin{align*}
\frac{1}{1 + \gamma t} \leq  \
&\lambda_{n} \leq \frac{1}{1 + \gamma b} 
\numberthis \label{scale-bound} \\
\end{align*}
\end{lemma}
}

\newcommand{\ProofLambda}{
\begin{proof}
From Lemma~\ref{lemma:gradient} we have
\begin{align*}
\ell' (y_{n}; x_{n}^\top  \theta_{n}) = \lambda_{n} \ell' (y_{n}; x_{n}^\top  \theta_{n-1}),
\numberthis \label{scaled-const-proof-1}
\end{align*}

where the derivative of $\ell$ is with respect to the natural parameter $x ^\top  \theta$. Using the definition of the implicit update Equation~\eqref{eq:implicit},
\begin{align*}
\theta_{n} = \theta_{n-1} + \gamma \lambda_{n} \ell ' (y_{n}; x_{n}^\top  \theta_{n-1}) x_{n}.
\numberthis \label{scaled-const-proof-2}
\end{align*}

We substitute this definition of $\theta_{n}$ into Equation\eqref{scaled-const-proof-1} and perform a Taylor approximation on $\ell ' $. Recall Taylor approximation for a function $f$, $f (x) = f (a) + f' (\xi) (x - a)$ where $\xi$ lies in the closed interval between $a$ and $x$. From Equation\eqref{scaled-const-proof-2} we let $\theta_{n-1} = a$ and $\gamma \lambda_{n} \ell ' (y_{n}; x_{n}^\top  \theta_{n-1}) x_{n} = (x - a)$. Also, by the Chain rule $\frac{\delta}{\delta \theta} \ell' (y; x ^\top  \theta) = \ell'' (y ; x ^\top  \theta) x^\top  $. Thus we obtain,
\begin{align*}
\ell' (y_{n}; x_{n}^\top  \theta_{n})
&= \ell' (y_{n}; x_{n}^\top  \theta_{n-1})
+ \ell'' (y_{n} ; x_{n} ^\top  \tilde{\theta}) x_{n}^\top  \cdot \gamma \lambda_{n} \ell ' (y_{n}; x_{n}^\top  \theta_{n-1}) x_{n} \\
&= \ell' (y_{n}; x_{n}^\top  \theta_{n-1})
+ \gamma \lambda_{n} \ell'' (y_{n} ; x_{n} ^\top  \tilde{\theta}) \ell ' (y_{n}; x_{n}^\top  \theta_{n-1}) x_{n}^\top  x_{n}
\numberthis \label{scaled-const-proof-3}
\end{align*}

where $\tilde{ \theta } = \delta \theta_{n-1} + (1 - \delta) \theta_{n}$ and $\delta \in [0,1]$. 

By combining Equation\eqref{scaled-const-proof-1} with Equation\eqref{scaled-const-proof-3} and cancelling out the first derivative term we get
\begin{align*}
\lambda_{n} 
&= 1 + \gamma \lambda_{n} \ell'' (y_{n} ; x_{n} ^\top  \tilde{\theta}) x_{n}^\top  x_{n} \\
\lambda_{n} (1 - \gamma \ell '' ( y_{n} ; x_{n} ^\top  \tilde{\theta} ) \| x \|^{2} ) 
&= 1 \\
\left( 1 + \gamma \ \text{trace} ( \hat{ \mathcal{I} }_{n} ( \tilde{ \theta } ) ) \right) \lambda_{n} 
&\leq 1
\ [ \textit{where } \hat{ \mathcal{I} } \textit{ is the observed Fisher information} ] \numberthis \label{scaled-const-proof-4}\\
(1 + \gamma b) \lambda_{n} 
&\leq 1 
\ [ \textit{By Assumption 1 (b)}] 
\numberthis \label{scaled-const-proof-5} \\
\textit{Now we get the other bound,} \\
(1 + \gamma t) \lambda_n &\geq 1 
\ [ \textit{By Assumption 1 (b)}] 
\end{align*}
\end{proof}
}

\newcommand{\TheoremPflugFiniteShort}{
\begin{theorem}
\label{theorem:pflugFinite}
Under certain assumptions, the 
convergence diagnostic in Algorithm~\ref{algo:pflug} for 
constant rate SGD procedure in Equation~\eqref{eq:sgd} satisfies 
$\Ex{S_n - S_{n-1}} < 0$ as $n\to\infty$, and so
the algorithm terminates almost surely.
\end{theorem}
}

\newcommand{\TheoremBound}{
\begin{theorem}
\label{theorem:isgd}
Suppose that Assumptions 1(a) - (e) hold. Then, 
\begin{align}
\Ex{||\theta_n - \theta_\star||^2} \leq& 
\left(1 -  \frac{ 2 \gamma \lambda }{ 1 + \gamma t } + \frac{2 \gamma^{2} L^{2}}{(1 + \gamma b)^{2}} \right)^{n}
\Ex{||\theta_{n-1} - \theta_\star||^2} \\
& \ + \frac{ \gamma \sigma^{2} (1 + \gamma t) }{ \lambda (1 + \gamma b)^2 - \gamma L^{2} (1 + \gamma t) }
\label{bound}
\end{align}
\end{theorem}
}

\newcommand{\ProofBound}{
\begin{proof}
Starting from the implicit update \eqref{eq:implicit}, we have 
\begin{align*}
\theta_{n} - \theta_{*} =& \ \theta_{n-1} - \theta_{*} + \gamma \nabla \ell (y_{n} ; x_{n}^\top  \theta_{n}) \\
\theta_{n} - \theta_{*} =& \ \theta_{n-1} - \theta_{*} + \gamma \lambda_{n} \nabla \ell ( y_{n}; x_{n}^\top  \theta_{n-1} )
\ [ \textit{By Lemma~\ref{lemma:gradient}} ] \\
\| \theta_{n} - \theta_{*} \|^{2} =& \ \| \theta_{n-1} - \theta_{*} \|^{2}  \\
&+ 2 \gamma \lambda_{n} ( \theta_{n-1} - \theta_{*} )^\top  \nabla \ell ( y_{n}; x_{n}^\top  \theta_{n-1} ) \\
&+ \| \gamma \lambda_{n} \nabla \ell ( y_{n}; x_{n}^\top  \theta_{n-1} ) \|^{2} 
\numberthis \label{expansion}
\end{align*}

To bound the last term,
\begin{align*}
\| \gamma \lambda_{n} &\nabla \ell ( y_{n}; x_{n}^\top  \theta_{n-1} ) \|^{2} \\
&= \gamma^{2} \lambda_{n}^{2}
\| \nabla \ell ( y_{n}; x_{n}^\top  \theta_{n-1} ) \|^{2} \\
&= \gamma^{2} \lambda_{n}^{2}
\| \nabla \ell ( y_{n}; x_{n}^\top  \theta_{n-1} ) 
- \nabla \ell ( y_{n}; x_{n}^\top  \theta_{*} ) + \nabla \ell ( y_{n}; x_{n}^\top  \theta_{*} )  \|^2 \\
&\leq 2 \gamma^{2} \lambda_{n}^{2}
\| \nabla \ell ( y_{n}; x_{n}^\top  \theta_{n-1} ) - \nabla \ell ( y_{n}; x_{n}^\top  \theta_{*} )  \|^{2}
+ 2 \gamma^{2} \lambda_{n}^{2}
\| \nabla \ell ( y_{n}; x_{n}^\top  \theta_{*} )  \|^2 \\
&\leq 2 \left( \frac{\gamma}{1 + \gamma b} \right)^{2}
\left( \| \nabla \ell ( y_{n}; x_{n}^\top  \theta_{n-1} ) - \nabla \ell ( y_{n}; x_{n}^\top  \theta_{*} ) \|^{2}
+ \| \nabla \ell ( y_{n}; x_{n}^\top  \theta_{*} ) \| ^2\right) \\
& \quad [ \textit{By Lemma \ref{lemma:gradBd}} ]
\numberthis \label{last-1}
\end{align*}

Taking expectation of both sides of Equation\eqref{last-1},
\begin{align*}
\mathbb{E} ( \| \gamma \lambda_{n} &\nabla \ell ( y_{n}; x_{n}^\top  \theta_{n-1} ) \|^{2} ) \\
&\leq 2 \left( \frac{\gamma}{1 + \gamma b} \right)^{2}
\left[ \mathbb{E} ( \| \nabla \ell ( y_{n}; x_{n}^\top  \theta_{n-1} ) - \nabla \ell ( y_{n}; x_{n}^\top  \theta_{*} ) \|^{2} )
+ \mathbb{E} (\| \nabla \ell ( y_{n}; x_{n}^\top  \theta_{*} ) \|^{2}) \right] \\
&\leq 2 \left( \frac{\gamma}{1 + \gamma b} \right)^{2} 
\left( L^{2} \| \theta_{n-1} - \theta_{*} \|^{2} + \sigma^{2} \right)
\ [ \textit{ By Lipschitz and gradient bound, Assumption 1 (c), (d) } ]
\numberthis \label{last-2}
\end{align*}

We can bound the expectation of the second term as 
\begin{align*}
\mathbb{E} ( 2 \lambda_{n} & \gamma ( \theta_{n-1} - \theta_{*} )^\top  \nabla \ell ( y_{n}; x_{n}^\top  \theta_{n-1} ) ) \\
&\geq \frac{ 2 \gamma }{ 1 + \gamma t } 
\mathbb{E} \left( ( \theta_{n-1} - \theta_{*} )^\top  \nabla \ell ( y_{n}; x_{n}^\top  \theta_{n-1} ) \right) 
\ [ \textit{By Lemma \ref{lemma:gradBd} } ] \\
&\geq  \frac{ 2 \gamma }{ 1 + \gamma t } 
\mathbb{E} \left( ( \theta_{n-1} - \theta_{*} )^\top  \nabla h ( \theta_{n-1} ) \right)
\ [where \ \nabla h ( \theta_{n-1} ) = \mathbb{E} ( \nabla \ell ( y_{n}; x_{n}^\top  \theta_{n-1} ) | \mathcal{F}_{n-1} ) ] \\
&\leq - \frac{ 2 \gamma \lambda }{ 1 + \gamma t } 
\mathbb{E} ( \| \theta_{n-1} - \theta_{*} \|^{2} )
\ [ \textit{By strong convexity, Assumption 1 (b) } ]
\numberthis \label{second}
\end{align*}

Taking expectations in \eqref{expansion} and substituting inequalities \eqref{last-2} and \eqref{second} into \eqref{expansion}, and again taking expectation, yields the recursion,
\begin{align*}
\mathbb{E} ( \| \theta_{n} - \theta_{*} \|^{2} ) 
\leq \left(1 -  \frac{ 2 \gamma \lambda }{ 1 + \gamma t } + \frac{2 \gamma^{2} L^{2}}{(1 + \gamma b)^{2}} \right)
\mathbb{E}( \| \theta_{n-1} - \theta_{*} \|^{2} ) 
+ 2 \left( \frac{\gamma \sigma}{1 + \gamma b} \right)^{2}
\numberthis \label{recursion1}
\end{align*}

Let $\delta_{n} \equiv \mathbb{E} ( \| \theta_{n} - \theta_{*} \|^{2} )$. 
We can now derive the bound of the theorem as follows:
\begin{align*}
\delta_{n}
&\leq \left(1 -  \frac{ 2 \gamma \lambda }{ 1 + \gamma t } + \frac{2 \gamma^{2} L^{2}}{(1 + \gamma b)^{2}} \right)^{n} \delta_{0}
+ \sum_{k=1}^{\infty} 
2 \left( \frac{\gamma \sigma}{1 + \gamma b } \right)^{2} 
\cdot  \left(1 -  \frac{ 2 \gamma \lambda }{ 1 + \gamma t } + \frac{2 \gamma^{2} L^{2}}{(1 + \gamma b)^{2}} \right)^{k} \\
& = \left(1 -  \frac{ 2 \gamma \lambda }{ 1 + \gamma t } + \frac{2 \gamma^{2} L^{2}}{(1 + \gamma b)^{2}} \right)^{n} \delta_{0}
+ 2 \left( \frac{\gamma \sigma}{1 + \gamma b} \right)^{2} 
\cdot \left(\frac{ 2 \gamma \lambda }{ 1 + \gamma t } - \frac{2 \gamma^{2} L^{2}}{(1 + \gamma b)^{2}} \right)^{-1} \\
& = \left(1 -  \frac{ 2 \gamma \lambda }{ 1 + \gamma t } + \frac{2 \gamma^{2} L^{2}}{(1 + \gamma b)^{2}} \right)^{n} \delta_{0}
+ \frac{ \gamma \sigma^{2} (1 + \gamma t) }{ \lambda (1 + \gamma b)^2 - \gamma L^{2} (1 + \gamma t) }
\end{align*}
\end{proof}
}

\newcommand{\LemmaViableLR}{
\begin{lemma}
\label{lemma:viableLR}
Suppose that Assumption 1(e) holds. The discount factor of the non-asymptotic bound in Theorem \ref{theorem:isgd} will be bounded $0 < \cdot < 1$ for all $\gamma >0$, and thus the mean squared error $\mathbb{E} ( \| \theta_{n} - \theta_{*} \|^{2} )$ will contract for all possible values of $\gamma$. In addition the stationary term will be $> 0$ for all $\gamma > 0$.
\end{lemma}
}

\newcommand{\ProofViableLR}{
\begin{proof}
The discount factor is bounded below by $\left(1 -  \frac{ 2 \gamma \lambda }{ 1 + \gamma b } + \frac{2 \gamma^{2} L^{2}}{(1 + \gamma b)^{2}} \right)$ because $b \leq t$. We will show that this term is bounded below by 0.

A quick manipulation of the algebra gives us
\begin{align*}
(lower \ bound) \quad
2 \gamma \lambda (1 + \gamma b) - 2 \gamma^{2} L^{2}
&< (1 + \gamma b)^{2} 
\numberthis \label{gamma-lower} \\
(upper \ bound) \quad
\gamma L^{2} (1 + \gamma t)
&< \lambda (1 + \gamma b)^2
\numberthis \label{gamma-upper} \\
(stationary \ bound) \quad
\gamma L^{2} (1 + \gamma t)
&< \lambda (1 + \gamma b)^2 
\numberthis \label{gamma-transient}
\end{align*}

Both the upper bound and stationary bound are satisfied by Assumption 1 (e). Further manipulating the lower bound, from Equation\eqref{gamma-lower},
\begin{align*}
2 \gamma \lambda + 2 \gamma^{2} \lambda b - 2 \gamma^{2} L^{2}
&< 1 + 2 \gamma b + \gamma^{2} b^{2} \\
\gamma^{2} (b^{2} - 2 \lambda b + 2 L^{2}) + \gamma (2 b - 2 \lambda) + 1 
&> 0
\numberthis \label{gamma-lower-2}
\end{align*}

Solving the equality of Equation\eqref{gamma-lower-2} (with the quadratic equation) gives us
\begin{align*}
&\frac{(2 \lambda - 2 b) \pm \sqrt{(2 b - 2 \lambda)^{2} - 4 (b^{2} - 2 \lambda b + 2 L^{2})} }{2 (b^{2} - 2 \lambda b + 2 L^{2})} \\
&= \frac{(2 \lambda - 2 b) \pm \sqrt{(4 b^{2} - 8 \lambda b + 4 \lambda^{2}) - 4 b^{2} + 8 \lambda b - 8 L^{2}} }{2 (b^{2} - 2 \lambda b + 2 L^{2})} \\
&= \frac{(2 \lambda - 2 b) \pm \sqrt{4 \lambda^{2} - 8 L^{2}}}{2 (b^{2} - 2 \lambda b + 2 L^{2})} \\
&= \frac{(\lambda - b) \pm \sqrt{\lambda^{2} - 2 L^{2}} }{(b^{2} - 2 \lambda b + 2 L^{2})}
\end{align*}

Recall that for a second-degree polynomial of the form $a_{2} x^{2} + a_{1} x + 1$, the convexity is determined by $a_{2}$. Because $L \geq \lambda$ (a standard assumption), the discriminant $(\lambda^{2} - 2 L^{2}) < 0$ and thus there are no real roots. Looking at the convexity,
\begin{align*}
(b^{2} - 2 \lambda b + 2 L^{2})
> (b^{2} - 2 \lambda b + \lambda^{2}) 
= (b - \lambda)^{2}
> 0
\end{align*}

The strict inequality is because of the following. For all observed Fisher information matrices, (with $p$ the dimesnion)
\begin{align*}
trace(\hat{ \mathcal{I} }_{n} (\theta)) &\geq b 
\Rightarrow 
\mathbb{E} trace(\hat{ \mathcal{I} }_{n} (\theta)) \geq b
\Rightarrow
\lambda \cdot p \geq b
\end{align*}

Thus for all $\gamma \in \mathbb{R}$ the lower bound represented by Equation\eqref{gamma-lower} is satisfied. We have zero real roots and a convex function.
\end{proof}

}

\newcommand{\lmin}{\mu}
\newcommand{\lmax}{L}
\newcommand{\pflug}{{\tt Pflug}}

\title{\bf \paperTitle}
\author[1]{Jerry Chee}
\affil[1]{University of Chicago}
\author[2]{Panos Toulis}
\affil[2]{University of Chicago, Booth School of Business}

\date{\today}

\begin{document}

\maketitle


\begin{abstract}
Many iterative procedures in stochastic optimization exhibit a transient phase followed by a stationary phase.~During the transient phase the procedure converges towards a region of interest, and during the stationary phase the procedure oscillates in that region, commonly around a single point. In this paper, we develop a statistical diagnostic test to detect such phase transition in the context of stochastic gradient descent with constant learning rate.~We present theory and experiments suggesting that the region where the proposed diagnostic is activated coincides with the convergence region.~For a class of loss functions, we derive a closed-form solution describing such region.~Finally, we suggest an application to speed up convergence of stochastic gradient descent by halving the learning rate each time stationarity is detected. This leads to a new variant of 
stochastic gradient descent, which in many settings is comparable to state-of-art.
\end{abstract}

\doublespacing

\section{Introduction}
We consider a classical problem in stochastic optimization stated as\begin{align}
\label{eq:star}
\theta_\star = \arg\min_{\theta\in\Theta} \Ex{\ell(y, x^\top \theta)},
\end{align}
where $\ell$ is the loss function, $y\in\mathbb{R}$ denotes the response, $x\in\mathbb{R}^p$ are the features, and $\theta$ are parameters in $\Theta\subseteq\mathbb{R}^p$. For example, the quadratic loss function is defined as $\ell(y, x^\top \theta) = (1/2)(y - x^\top \theta)^2$.
In estimation problems we typically have a finite data set $\{(x_i, y_i)\}$, $i=1, 2, \ldots, N$, from which we wish to estimate $\theta_\star$ by solving the empirical version of Equation~\eqref{eq:star}:
$$
\hat\theta = \arg\min_{\theta\in\Theta} \frac{1}{N}\sum_{i=1}^N 
\ell(y_i, x_i^\top \theta).
$$

When data size, $N$, and parameter size, $p$, are large classical methods for computing $\hat\theta$ fail. Stochastic gradient descent (SGD) is a powerful alternative~\citep{bottou2010large, bottou2012stochastic, toulis2015scalable, zhang2004solving} because it solves the problem in an iterative fashion through the procedure:
\begin{align}
\label{eq:sgd}
\theta_n  = \theta_{n-1}  - \gamma \nabla \ell(y_n, x_n^\top \theta_{n-1}).
\end{align}
Here, $\theta_{n-1}$ is the estimate of $\theta_\star$ prior to the $n$th iteration, $(x_n, y_n)$ is a random sample from the data, and $\nabla \ell$ is  the gradient of the loss with respect to $\theta$. 
Classical stochastic approximation theory~\citep{benveniste1990adaptive, borkar2008stochastic, robbins1951stochastic} suggests that SGD converges to a value $\theta_\infty$ such that $\Ex{\nabla \ell(y, x^\top \theta_\infty)}=0$, which 
under typical regularity conditions is equal to $\theta_\star$ when $N$ is infinite (streaming setting), or is equal to $\hat\theta$ when $N$ is finite. Going forward we 
assume the streaming setting for simplicity, but our results hold for 
finite $N$ as well.

Typically, stochastic iterative procedures start from some starting point and then move through a transient phase and towards a stationary phase~\citep{murata1998statistical}. 
In stochastic gradient descent this behavior is largely governed by parameter $\gamma > 0$, which is known as the learning rate, and can either be decreasing over $n$ (e.g., $\propto 1/n$), or constant. 
In the decreasing rate case, the transient phase is usually long, and can be impractically so if the rate is slightly misspecified~\citep{nemirovski2009robust, toulis2017asymptotic}, whereas the stationary phase involves SGD converging in quadratic mean to $\theta_\star$. 
When $\gamma$ is constant the transient phase is much shorter and less sensitive to the learning rate, whereas the stationary phase involves SGD oscillating within a region that contains $\theta_\star$.~In this paper, we focus on statistical convergence diagnostics for constant rate SGD because 
in this setting  a convergence diagnostic can be utilized to identify when there is no benefit in running the procedure longer. 

\subsection{Related work and contributions}
The idea that SGD methods are composed of a transient phase and a stationary phase (also known as search phase and convergence phase, respectively), has been expressed before~\citep{murata1998statistical}. However, 
 no principled statistical methods have been developed to address stationarity issues, and thereby guide empirical practice of SGD.~Currently, guidance is based on heuristics originating from optimization theory that aim to evaluate  the magnitude of SGD updates. For example, a popular method is to stop when $||\theta_n-\theta_{n-1}||$ is small according to some threshold, or when updates of the loss function have reached machine precision~\citep{bottou2016optimization, ermoliev1988numerical}.
 These methods, however, do not take into account the sampling variation in SGD estimates, and are  suited for deterministic procedures but not stochastic ones.

A more statistically motivated approach is to monitor the test error of SGD iterates on a hold-out validation test, concurrently with the main SGD iteration~\citep{blum1999beating, bottou2012stochastic}. 
One idea here is to stop the procedure when the validation error starts increasing. An important problem with this approach is that the validation error is also a stochastic process, and estimating when it actually starts increasing presents similar, if not greater, challenges to the original problem of detecting convergence to stationary phase. Furthermore, cross validation can be computationally costly in large data sets.
 
In stochastic approximation, methods to detect stationarity can be traced back to classical theory of stopping times~\citep{pflug1990non, yin1989stopping}.~One important method, which forms the basis of this paper, is Pflug's procedure~\citep{pflug1990non} that keeps a running average of the inner product of successive gradients $\nabla_{n-1} \ell^\top
\nabla_n\ell$, where we defined $\nabla_j \ell = \nabla \ell(y_j, x_j^\top \theta_{j-1})$. 
 The underlying idea is that in the transient phase the stochastic gradients point roughly to the same direction, and thus their inner product is positive. In the stationary phase, SGD with constant rate moves haphazardly around $\theta_\star$, and so the gradients point to different directions making the inner product negative. 

The intuition that a negative inner product of successive gradients indicates convergence underlies accelerated methods in stochastic approximation~\citep{delyon1993accel, kesten1958accel, roux2009accel}. The accelerated methods, however, take this intuition as a given, whereas we develop theory for it to define a formal convergence testing procedure.
Recently, another related idea is that of gradient diversity~\citep{yingradient}, which is used to understand  why speedup gains in batch SGD saturate with increasing batch size. An important difference is that gradient diversity calculates the inner products at a fixed parameter value $\theta$, 
whereas stochastic approximation methods, including this paper, use successive parameter values.

\subsubsection{Overview of results and contributions}
Our contributions in this paper can be summarized as follows.
In Section~\ref{section:diagnostic}, we develop a formal convergence diagnostic test for SGD, which combines Pflug's stopping time procedure~\citep{pflug1990non} with SGD in Equation~\eqref{eq:sgd} to detect when SGD exits the transient phase and enters the stationary phase.~We note that by convergence of SGD with constant rate we do not mean convergence to a single point but convergence to the stationarity region.~We prove a general result that the diagnostic indeed is activated almost surely.
We illustrate through an example, where conditional on the diagnostic 
being activated, the distance $||\theta_n- \theta_\star||$ is uncorrelated with 
the starting distance $||\theta_0-\theta_\star||$, implying that 
the diagnostic captures the transition from transient to stationary phase.
In Section~\ref{section:linear}, we develop theory for quadratic loss, and derive a closed-form solution describing the region where the diagnostic is activated.~In Section~\ref{section:ext_glm}, we present extensions beyond the quadratic loss.~In Section~\ref{section:ext_sgdhalf} we suggest an application of the diagnostic in speeding up SGD by halving the learning rate each time convergence is detected. This leads to a new SGD procedure, named \SGDhalf, which is comparable to state-of-art procedures, such as variance-reduced SGD~\citep{johnson2013accelerating} and averaged SGD~\citep{bottou2010large, xu2011towards}, in Sections~\ref{section:ext_experiment_sim} and~\ref{section:ext_experiment_bench}.

\section{Convergence diagnostic}
\label{section:diagnostic}
Before we develop the formal diagnostic, we present theory that supports the existence of a transient and stationary phase of SGD. The theory suggests that the mean squared error of SGD  has a bias term from distance to the starting point, and a variance term from noise in stochastic  gradients.
\begin{theorem}
\label{theorem:sgd}
 [\citep{moulines2011non, needell2014stochastic}]
Under certain assumptions on the loss function, there are positive constants 
$A_\gamma, B$ such that, for every $n$,  it holds that
$$
\Ex{||\theta_n - \theta_\star||^2} \le \Ex{||\theta_0-\theta_\star||^2} e^{-A_\gamma n}  + B \gamma.
$$
\end{theorem}

{\em Remarks.} The constants $A_\gamma, B$ differ depending on the analysis. For example, Bach and Moulines~\citep{moulines2011non} use $A_\gamma \approx \gamma\lmin/4 - \gamma^2\lmax^2$, 
where $\lmin$ is the strong convexity constant of expected loss
$f(\theta) = \Ex{\ell(y, x^\top \theta) | \theta}$, and $\lmax$ is the Lipschitz constant of $\nabla \log\ell(y, x^\top \theta)$; and $B =\sigma^2/\mu$, where $\sigma^2$ is an upper bound for the variance of $||\nabla\log \ell(y, x^\top \theta_\star)||^2$.
Needell and Srebro~\citep{needell2014stochastic} use $A_\gamma \approx 2\gamma\lmin- 2\gamma^2\lmin\lmax$ and 
$B = \sigma^2/ \lmin(1-\gamma \lmax)$.

Despite such differences, all analyses suggest that the SGD procedure with constant rate goes through a transient phase exponentially fast during which it forgets the initial conditions 
$\Ex{||\theta_0-\theta_\star||^2}$, and then enters a stationary phase during which it oscillates around $\theta_\star$, roughly at a region of radius 
 $R_\gamma = O(\sqrt\gamma)$. A trade-off exists here: reducing $\gamma$ will make the oscillation radius, $R_\gamma$, smaller but escaping the transient phase becomes much slower; for instance, in the extreme case where $\gamma=0$ the procedure will never exit the transient phase.

Despite the theoretical insights it offers, Theorem~\ref{theorem:sgd} has limited practical utility for estimating the phase transition in SGD. One approach could be to find the value of $n$ for which
$\Ex{||\theta_0-\theta_\star||^2} e^{-A_\gamma n} = 0.01 B \gamma$,
that is, the initial conditions have been discounted to 1\% of the 
stationary variance.
That, however, requires estimating $\lmin, \lmax$, $\sigma^2$, and $\Ex{||\theta_0-\theta_\star||^2}$, which is challenging. 
In the following section, we develop a concrete statistical diagnostic to estimate the phase transition and detect convergence of SGD in a much simpler way.

\subsection{{\tt Pflug} diagnostic}
In this section, we develop a convergence diagnostic for SGD procedures that relies on Pflug's procedure~\citep{ pflug1992gradient} in stochastic approximation. The diagnostic is presented as Algorithm~\ref{algo:pflug} and concrete instances under quadratic loss along with theoretical analysis are presented in Section~\ref{section:linear}, 
with extensions in Section~\ref{section:extensions}.

The diagnostic is defined by a random variable $S_n$ that keeps the running sum of the inner product of successive stochastic gradients, as shown in Line 7. 
The idea is that in the transient phase SGD moves  towards $\theta_\star$ by discarding initial conditions, and so the  stochastic gradients point to the same direction, on average.
This implies that the inner product of successive stochastic gradients is likely positive in the transient phase. In the stationary phase, however, SGD is oscillating around $\theta_\star$ at a distance bounded by Theorem~\ref{theorem:sgd}, and so the gradients point to different directions.
This implies a negative inner product on average during the stationary phase. When the statistic $S_n$ changes sign from positive to negative, this is a good signal that the procedure has exited the transient phase. 

Since our convergence diagnostic is iterative we need to show that it eventually terminates with an answer. 
In Theorem~\ref{theorem:pflugFinite} that follows we prove that $\Ex{S_n - S_{n-1}} < 0$ as $n\to\infty$, and so Algorithm~\ref{algo:pflug} indeed 
terminates almost surely. For brevity, we state the theorem without technical details. The full assumptions and proof can be found in the supplementary material.

\begin{algorithm}[t!]
\setstretch{1.35}

\caption{
\setstretch{1.15}
{\tt Pflug} diagnostic for convergence of stochastic gradient descent.\newline {\bf Input}: starting point $\theta_0$, data $\{(y_1, x_1), (y_2, x_2), \ldots \}$, 
$\gamma > 0$, {\tt burnin > 0}.
\newline {\bf Output}: Iteration when SGD in Equation~\eqref{eq:sgd} is estimated to enter stationary phase.}
\begin{algorithmic}[1]
	\STATE $S_0 \gets 0$
	\STATE $\theta_1 \gets \theta_0 - \gamma \nabla \ell(y_1, x_1^\top \theta_0)$
	\FORALL{$n \in \{2, 3,\cdots\}$}
	\STATE Sample $(x_n, y_n)$
	\STATE Define $\nabla\ell_n = \nabla\ell(y_n, x_n^\top \theta_{n-1})$.
	\STATE $\theta_n \gets \theta_{n-1} - \gamma \nabla\ell_n$.
	\STATE $S_n \gets S_{n-1} + \nabla \ell_n^\top \nabla\ell_{n-1}$.
	\IF {$n  > {\tt burnin}$ and $S_n  < 0$}
	\RETURN $n$
	\ENDIF
	\ENDFOR
\end{algorithmic}
\label{algo:pflug}
\end{algorithm}

\TheoremPflugFiniteShort

{\em Remarks.}  Theorem~\ref{theorem:pflugFinite} shows that the inner product of successive gradients is negative in expectation as the iteration number increases. Roughly speaking, 
when $\theta_n$ is very close to $\theta_\star$ the dominant force is the variance in the 
stochastic gradient pulling the next iterates away from $\theta_\star$;
when $\theta_n$ is far from $\theta_\star$ the dominant force is the bias in the stochastic gradient, which instead pulls the next iterates towards $\theta_\star$. This implies that the running sum of successive gradients will eventually become negative at a finite iteration number, and so by the law of large numbers the  diagnostic returns a value almost surely. 

\section{Quadratic loss model}
\label{section:linear}
In this section, we attempt to gain analytical insight into our convergence diagnostic of Algorithm~\ref{algo:pflug} by assuming simple quadratic loss function, i.e., $\ell(y, x^\top \theta) = (1/2) (y-x^\top \theta)^2$ 
and $\nabla\ell(y, x^\top \theta) = -(y-x^\top \theta) x$.
%
Consider the case where $\theta_{0} = \theta_\star$, i.e., the procedure 
starts in the stationary region. Let $y_n = x_n^\top \theta_\star + \varepsilon_n$, where $\varepsilon_n$ are zero-mean random variables, $\Ex{\varepsilon_n | x_n}=0$.
Then, 
\begin{align}
\theta_{1} & = \theta_\star + \gamma (y_1 - x_1^\top\theta_\star)x_1 = \theta_\star + \gamma \varepsilon_1 x_1,\nonumber
\end{align}
from which it follows that
\begin{align}
\label{eq:sn}
S_2 - S_{1} & = (y_2-x_2^\top \theta_1)(y_1 - x_1^\top\theta_0) x_2^\top x_{1}\nonumber = (\varepsilon_2 - \gamma \varepsilon_1 x_2^\top x_1)
     \varepsilon_1 x_2^\top x_1.\nonumber\\
\Ex{S_2 - S_1} & = -\gamma \Ex{\varepsilon_1^2}\Ex{(x_2^\top x_1)^2} < 0.
\end{align}
Thus, when the procedure starts at true parameter value, $\theta_\star$, 
the diagnostic is decreased in expectation, and eventually at some iteration $\tau$ the statistic $S_\tau$ becomes negative and the diagnostic is activated at  $\tau$.
We generalize this result in the following theorem.

\LinearTheorem

{\em Remarks.} 
Theorem~\ref{theorem:linear} shows that the boundary 
surface that separates the two regions where the 
test statistic $S_n$ increases or decreases in expectation looks like an ellipse, 
for large enough $\gamma$. Regardless of the choice of $\gamma$, when $\theta_n$ is close enough to $\theta_\star$, the diagnostic is guaranteed to decrease in expectation since the only remaining term is 
$-\gamma c^2\sigma^2 < 0$.

The result also shows the various competing forces between bias and variance in the stochastic gradients as they relate to how the diagnostic behaves.
For instance, when $\theta_n$ is very close $\theta_\star$, 
larger $\sigma^2$ (noise in stochastic gradient) contributes to a faster decrease of the diagnostic in expectation, but at the cost of higher variance.
The contribution of the other term, $c^2$, is less clear. 
For instance, $c$ is large when there is strong collinearity in features 
$x$, which may contribute to decreasing $S_n$. But strong collinearity 
also implies that $C$ is almost positive definite which contributes positive values to $S_n$, thus counteracting the contribution of $c$.
Note that $D$ is a positive definite matrix but $C$ may not be. This implies that careful selection of $\gamma$ may be necessary for the diagnostic to work well. For example, when $\gamma$ is very small and $C$ is positive definite, then $S_n$ 
will converge to a negative number slowly. One way to alleviate this sensitivity to the learning rate is through implicit updates~\citep{toulis2014statistical}, which we explore in the following section.

\subsection{Implicit update}
\label{section:linear_implicit}
As mentioned above the \pflug\ diagnostic is sensitive to the choice of learning rate $\gamma$.
When $\gamma$ is small and $C$ is positive definite, $S_n$ will be mostly increasing during the transient phase, which 
makes convergence slower. But choosing a large learning rate can easily lead to numerical instability. One way to alleviate such sensitivity to the learning rate is to use the SGD procedure with an implicit update as follows:
\begin{align}
\label{eq:implicit}
\theta_n  = \theta_{n-1} - \gamma \nabla\ell(y_n, x_n^\top \theta_n).
\end{align}
Note that $\theta_n$ appears on both sides of the equation. In the quadratic loss model we can solve exactly the implicit equation as follows:
\begin{align}
\label{eq:implicit_linear}
\theta_n  = (I + \gamma x_n x_n^\top)^{-1} (\theta_{n-1} + \gamma y_n x_n).
\end{align}
Implementing the procedure in Eq.~\eqref{eq:implicit_linear} is fast 
since it is equivalent to 
$\theta_n  =(\theta_{n-1} + \gamma y_n x_n) / (1+\gamma ||x_n||^2)$.
More generally, the implicit update in Equation~\ref{eq:implicit} can be computed efficiently in many settings through a one-dimensional root-finding procedure~\citep{toulis2014statistical}.
Previous work on implicit SGD (ISGD) has shown that 
implicit procedures have similar asymptotic properties with standard SGD procedures with numerical stability as an added benefit. 
Since most related work on ISGD methods is with respect to decreasing learning rate procedures~\citep{bertsekas2011incremental, kulis2010implicit, toulis2017asymptotic, toulis2014statistical}, we provide an analysis for constant rate ISGD as in Equation~(4) in the supplementary material.
We note that ISGD procedures are related to proximal updates in stochastic optimization~\citep{parikh2013proximal, rosasco2014convergence, xiao2014proximal}, but these methods differ  from ISGD methods in that they employ a combination of classical SGD with deterministic proximal operators, whereas ISGD's proximal operator is purely stochastic.

The following theorem shows that the implicit update in the linear model mitigates the sensitivity of the \pflug\ diagnostic to the choice of the learning rate. 

\LinearTheoremImplicit
{\em Remarks.}
Theorem~\ref{theorem:linear_implicit} shows that the diagnostic is more stable with the ISGD procedure than with the classical SGD procedure. By stability we mean two things.
First, even when classical SGD diverges the convergence diagnostic may still declare convergence. Consider, for example, the simple model $\theta_n = \theta_{n-1} + \gamma (y_n - \theta_{n-1})$, where $y \sim N(\theta_\star, 1)$. If $\gamma > 1$ the classical SGD procedure will diverge. However, the diagnostic will declare convergence almost immediately because by Theorem~\ref{theorem:linear} it decreases, in expectation, for every $\theta$. Such inconsistency due to instability of classical SGD cannot happen with implicit SGD.
 
Second, generally speaking, empirical performance of the diagnostic
under implicit SGD matches theory better than under classical SGD.
This is illustrated in the following section, where the region 
of diagnostic convergence is smooth and elliptical under implicit SGD, as predicted by Theorem~\ref{theorem:linear_implicit}; 
under classical SGD, the corresponding region does not follow 
Theorem~\ref{theorem:linear} as closely due to sensitivity to learning rate specification.

\subsection{Illustration}
\label{section:illustration}
\begin{figure}[t!]
\centering
  \includegraphics[width=.7\textwidth]{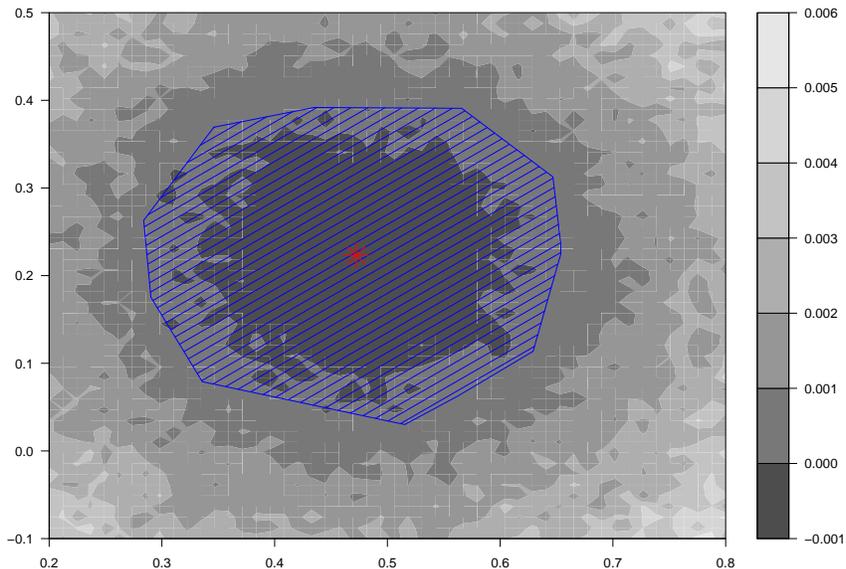}
  \caption{Shaded area in the center: region where \pflug\ diagnostic is 
  decreased in expectation. Polygon around shaded area: convergence region of SGD where iterates oscillate around (empirically calculated).
  Color legend on the right: values of expected increase (or decrease) 
  of the diagnostic.}
  \label{fig:pflug}
\end{figure}

Here, we illustrate the main results of Theorem~\ref{theorem:linear_implicit} through Figure~\ref{fig:pflug}, which can be described as follows.
The shaded areas in the figure show how
the \pflug\ diagnostic changes in expectation when the SGD iterate falls in the region. In other words, 
every point $\theta$ in the figure is shaded by the value 
$\Delta_n^{\mathrm{im}}(\theta)$, as defined in Theorem~\ref{theorem:linear_implicit}. 

Various shades of grey indicate the magnitude of change. 
The darkest-shaded region corresponds to the region where the diagnostic 
decreases in expectation, that is, $\Delta_n^{\mathrm{im}}(\theta) \le 0$ for all $\theta$ in that region. We call this the \pflug\ region. Note that the \pflug\ region is centered roughly around $\theta_\star$, the true parameter value.~Inside the \pflug\ region the diagnostic is decreased in expectation, and outside of the region it is increased. Furthermore, the expected change in the diagnostic is uniform in distance to the center of the \pflug\ region, which is roughly $\theta_\star$: the farther we move away from the center $\theta_\star$ the 
larger the expected increase of the diagnostic becomes.

The blue polygon shaded with diagonal lines corresponds to empirical estimations of the convergence region of SGD, 
defined as the region where SGD iterates have oscillated around for 95\% of the time calculated over 1000 simulations.
The polygon shows that the \pflug\ region 
approximates very well the actual convergence region of SGD. This is remarkable because the \pflug\ region can be calculated from data using the convergence diagnostic, whereas by Theorem~\ref{theorem:sgd} the SGD convergence region cannot be calculated without knowledge of $\theta_\star$ and other unknown parameters.

\subsection{Simulated example}
\label{section:linear_simulation}

Next, we test the {\tt Pflug} diagnostic through a simulated experiment. The experimental setup is as follows. We set $p=20$ as the parameter dimension, and also set $N=5000$ as the data set size and fix $\theta_\star\in\mathbb{R}^p$ with $\theta_{\star, j} = 10 e^{-0.75 j}$; this ensures some variation and sparsity in the parameter values.
We sample features as $x_i \sim \mathcal{N}_p(0, I)$, where $\mathcal{N}_p$ denotes a $p$-variate normal distribution, $I$ is the identity matrix, and $i=1, 2, \ldots N$.
We sample outcomes as  $y_i = x_i^\top  \theta_\star + \mathcal{N}(0, \sigma^2)$, where $\sigma=3$.

For given $\gamma$ we run Algorithm~\ref{algo:pflug} with {\tt burnin} = $0.1 N$, for various 
values of the starting point $\theta_0$ sampled as $\mathcal{N}_p(\theta_\star, \sigma_0^2 I)$, where $\sigma_0=2$. Let $\Le_n = ||\theta_n - \theta_\star||^2$, then 
for each run we store the tuple 
$$
(\gamma, \tau, \Le_0, \Le_{\tau/2}, \Le_{2\tau}),
$$
where $\tau$ is the output of Algorithm~\ref{algo:pflug}, i.e., 
the iteration at which the {\tt Pflug} diagnostic detected convergence.
The idea in this experimental evaluation is that if the convergence diagnostic detects convergence accurately, 
 iterates earlier than convergence, say, $\theta_{\tau/2}$, will depend on the initial conditions $\theta_0$ more than iterates later than convergence, say, $\theta_{2\tau}$. Thus, for given $\gamma$ and $\tau$, 
we should expect a much higher correlation between $\Le_{\tau/2}$ and $\Le_0$ than between $\Le_{2\tau}$ and $\Le_0$. 
To test this hypothesis,  for a given value of $\gamma$ we draw 100 independent samples of $(\Le_0, \Le_{\tau/2}, \Le_{2\tau})$.
With these samples we regress $\Le_{\tau/2}$ on $\Le_0$ and $\Le_{2\tau}$ on $\Le_0$ in two normal linear regression models. 
Table~\ref{table:diagnostic} summarizes the regression results from this experiment.
In the second and third column of Table~\ref{table:diagnostic} we report the regression coefficients of $\Le_0$ in the two model fits, respectively, and also report statistical significance.

\renewcommand{\arraystretch}{1.3}
\begin{table}[h!]
  \caption{Experimental evaluation of convergence diagnostic over 100 runs per learning rate value. Significance levels: *** = $<0.1\%$; ** = $1\%$; *  = $5\%$; . = $10\%$}
  \label{table:diagnostic}
  \centering
  \begin{tabular}{ccc}
    \toprule
 & $\Le_{\tau/2} = \beta_{\tau/2} \Le_0 + \varepsilon$ & $\Le_{2\tau} = \beta_{2\tau} \Le_0 + \varepsilon$ \\
   $\gamma$     & $\beta_{\tau/2}$     & $\beta_{2\tau}$ \\
    \midrule
    $0.02$ & $0.17$ **  &  $0.01$ .     \\
    $0.05$     & $0.20$ *** & $-0.008$     \\
    $0.1$     & $0.09$ **       & $-0.0007$  \\
    $0.2$ & $0.06$ ** & $0.005$ \\
    $0.5$ & $0.09$ ***   & $-0.008$ \\
    $1.0$ & $0.06$ * & $0.02$ * \\
    $ 2.0$ & $0.06$ ** & $0.009$\\
    $5.0$ & $0.07$ ** & $-0.012$\\
    \bottomrule
  \end{tabular}
\end{table} 

From the table, we see that the regression coefficient corresponding to $\Le_{\tau/2}$ is always positive and statistically significant, whereas the coefficient is mostly not significant for $\Le_{2\tau}$. This suggests that $\Le_{\tau/2}$ depends on initial conditions $\Le_0$, and thus stationarity has not yet been reached at iteration $\tau/2$. 
In contrast, $\Le_{2\tau}$ does not depend on initial conditions $\Le_0$, and thus stationarity has likely occurred after iteration $\tau$. 
This is evidence indicating that the {\tt Pflug} diagnostic performs reasonably well in estimating the switch of SGD from its transient phase to its stationary phase.

We note that in the regression evaluation we had to control for $\tau$ (by using it as a regressor) because the iteration number is correlated with mean-squared error (larger values for $\tau$ are correlated with smaller error).

\section{Extensions and applications}
\label{section:extensions}
In this section we consider extensions of the \pflug\ diagnostic to a more broad family of loss functions inspired by generalized linear models (GLMs). We also consider 
an application of the diagnostic to speed up convergence of SGD with constant rate.

\subsection{Generalized linear loss}
\label{section:extensions}
In this section we consider extensions of the \pflug\ diagnostic to a more broad family of loss functions inspired by generalized linear models (GLMs). We also consider an application of the diagnostic to speed up convergence of SGD with constant learning rate.

\subsection{Generalized linear loss}
\label{section:ext_glm}
Here, we consider the loss based on the GLM formulation~\citep{mccullagh1984generalized, toulis2014statistical} 
where $\ell(y, x^\top\theta) = -y \cdot x^\top\theta + f(x^\top\theta)$. 
For example, the quadratic loss is equivalent to $f(u) = u^2/2$. 
The logistic loss is when $y$ is binary and $f(u) = \log(1 + e^u)$.
In general, $f$ cannot be chosen arbitrarily---one standard choice is to define $f$ such that $e^{-\ell(y, x^\top\theta)}$ is a proper density, i.e., it integrates to one.
The following theorem generalizes the results in Section~\ref{section:linear} on the quadratic loss.

\TheoremGeneral 
{\em Remarks.} 
The result in Theorem~\ref{theorem:general} has the same structure as in Theorem~\ref{theorem:linear} so a direct analogy can be helpful. 
The terms $\sigma^2, c^2$ in the two theorems are identical, if we consider that for the quadratic loss it holds that $h(u) = u$.
The term $||C(\theta, \theta_\star)||^2$ in Theorem~\ref{theorem:general} corresponds to the term 
$(\theta-\theta_\star)^\top C (\theta - \theta_\star)$ in Theorem~\ref{theorem:linear}, and $D^2(\theta, \theta_\star)$ corresponds to
$(\theta-\theta_\star)^\top D (\theta - \theta_\star)$.
The terms are equal when we set $h(u)=u$, in which case $k=1$.
Thus, the diagnostic with the more general GLM loss has familiar properties. For example, when $\theta\approx\theta_\star$, i.e., when SGD is near the truth, $||C(\theta, \theta_\star)||^2\approx 0$ and 
$D^2(\theta, \theta_\star)\approx 0$, in which case the negative 
constant term dominates, and the test statistic decreases in expectation 
leading to activation of the diagnostic. 
One difference with the quadratic loss, however, is that as 
we move farther from $\theta_\star$ 
the statistic may change in a nonlinear way. 
Therefore the boundary separating the positive and negative regions 
of the diagnostic will generally not have the familiar smooth elliptical 
shape as in the quadratic loss (see Figure~\ref{fig:pflug}).
This may lead to more complex behavior for the diagnostic, which is open to future work.

Regarding the assumptions of Theorem~\ref{theorem:general}, we note that the constraint on derivative $h'$ is not particularly strict because in the GLM formulation $h'$ is guaranteed to be positive. The assumption is made to simplify the analysis, but 
can be improved by analyzing the quantity $h'(x^\top\theta_n)$ through existing analyses of $\theta_n$.

\subsection{\SGDhalf\ for fast convergence}
\label{section:ext_sgdhalf}
We now switch gears from analyzing the behavior of the \pflug\ diagnostic to using it in a practical application. Our suggested application
is to use the diagnostic within a SGD loop where the learning rate is halved and the procedure restarted each time convergence is detected.
We emphasize that our goal here is to illustrate the utility of our convergence diagnostic and 
not to exhaustively demonstrate the performance of the new procedure.
A full analysis of the proposed procedure is open to future work.

More specifically, the SGD procedure with constant rate has linear convergence to a stationary distance from $\theta_\star$ of $R_\gamma = O(\sqrt\gamma)$, as suggested by Theorem~\ref{theorem:sgd}.
It would therefore be beneficial to reduce the learning rate when we know that SGD iterates are oscillating around $\theta_\star$ in a ball of radius $R_\gamma$, so that 
the procedure moves to a ball with a smaller radius.
To implement such a procedure, however, would require knowing $\theta_\star$, and also knowing all parameters 
required to calculate $R_\gamma$. 
Our solution employs the \pflug\ convergence diagnostic to detect stationarity. Algorithm~\ref{algo:isgdHalf} describes such a procedure, named \SGDhalf, where the learning rate is halved upon detection of convergence (Line 10).

Note that implicit updates can be used in this algorithm as well; we call this modified algorithm \ISGDhalf. In our experiments in the following section, we employ \ISGDhalf\ because of the benefits in numerical stability from using implicit updates, as described earlier.

\begin{algorithm}[t!]
\setstretch{1.0}

\caption{
\setstretch{1.0}
Procedure~\SGDhalf.\newline {\bf Input}: $\theta_0$, data $\{(y_1, x_1), (y_2, x_2), \ldots \}$, 
$\gamma > 0$, {\tt burnin, maxit > 0}.
\newline {\bf Output}: Iteration $\tau > 0$, when SGD is estimated to have converged.}
\begin{algorithmic}[1]
	\STATE $s \gets 0$
	\STATE $\tau \gets 0$
	\STATE $\theta_1 \gets \theta_0 - \gamma \nabla \ell(y_1, x_1^\top \theta_0)$
		\FORALL{$n \in \{2, 3,\cdots\}$}
	\STATE $\theta_n \gets \theta_{n-1} - \gamma \nabla \ell(y_n, x_n^\top \theta_{n-1})$ 
	\STATE $s \gets  s + (\theta_n - \theta_{n-1})^\top  (\theta_{n-1} - \theta_{n-2})/ \gamma^2$
	\IF {$n  > \tau + {\tt burnin}$ and $s  < 0$}
	\STATE $\tau \gets n$
	\STATE $s\gets 0$
	\STATE $\gamma \gets \gamma/2$ 
		\IF {$\gamma < $ {\tt 1e-10} \AND $n > {\tt maxit}$}
	\RETURN $\theta_n$.
\ENDIF
	\ENDIF
	\ENDFOR
\end{algorithmic}
\label{algo:isgdHalf}
\end{algorithm}

\newcommand{\snr}{\text{SNR}}
\subsection{Simulated data experiments}
\label{section:ext_experiment_sim}
To evaluate the effectiveness of \ISGDhalf, we compare to other classical and state-of-the-art SGD methods. We first experiment on simulated data to better understand the performance of \ISGDhalf\ and its competition under various parameter settings. 
In particular, we compare the performance of procedure \ISGDhalf\  in Algorithm~\ref{algo:isgdHalf} against SVRG and classical ISGD on simulated data. The classical ISGD uses a learning rate of $O(1/n)$, which is optimized through pre-processing. The basic experimental setup is as follows. 

We consider settings of high and low signal to noise ration (SNR), and high and low dimension and test under the four combinations of these settings. For the high SNR case, we set $\snr=5$, where $\snr = \mathrm{trace}(Var(x)) /  p Var(y|x)$, and for the low SNR case we set $\snr=2$.
For the high dimension case we set $p = 150$ as the parameter dimension, and for the low dimension case we set $p = 10$. Given $p$, we fix $\theta_{*} \in \mathbb{R}^{p}$ such that $\theta_{\star, j} = 10 e^{-0.75 j}$. We set $N = 5000$ as the size of the data set. 
We sample features as $x_i \sim \mathcal{N}_p(0, I)$, where $i=1, 2, \ldots N$. We sample outcomes as  $y_i \sim \mathcal{N}(x_i^\top  \theta_\star , \sigma^2)$ for the normal model, and $y_i \sim \mathrm{Binom}(\exp(x_i^\top \theta_\star)/(1 + \exp(x_i^\top \theta_\star))$ for the logistic model, where $\mathrm{Binom(q)}$ denotes the binomial random variable with mean $q$.  The learning parameters for each SGD method were tuned to provide best performance through pre-processing.

\begin{figure*}[t]
\centering
\includegraphics[width=0.48\textwidth]{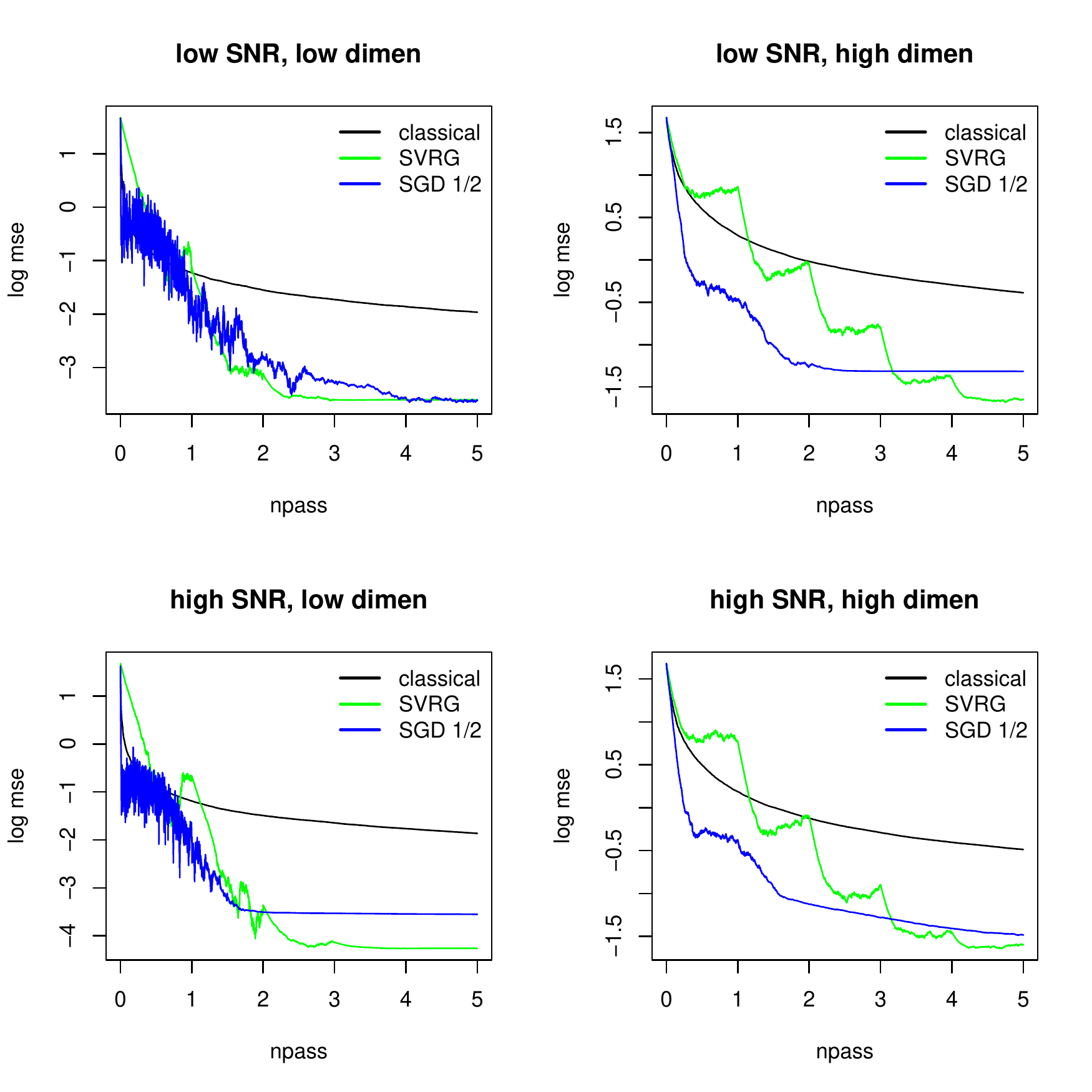}
\includegraphics[width=0.49\textwidth]{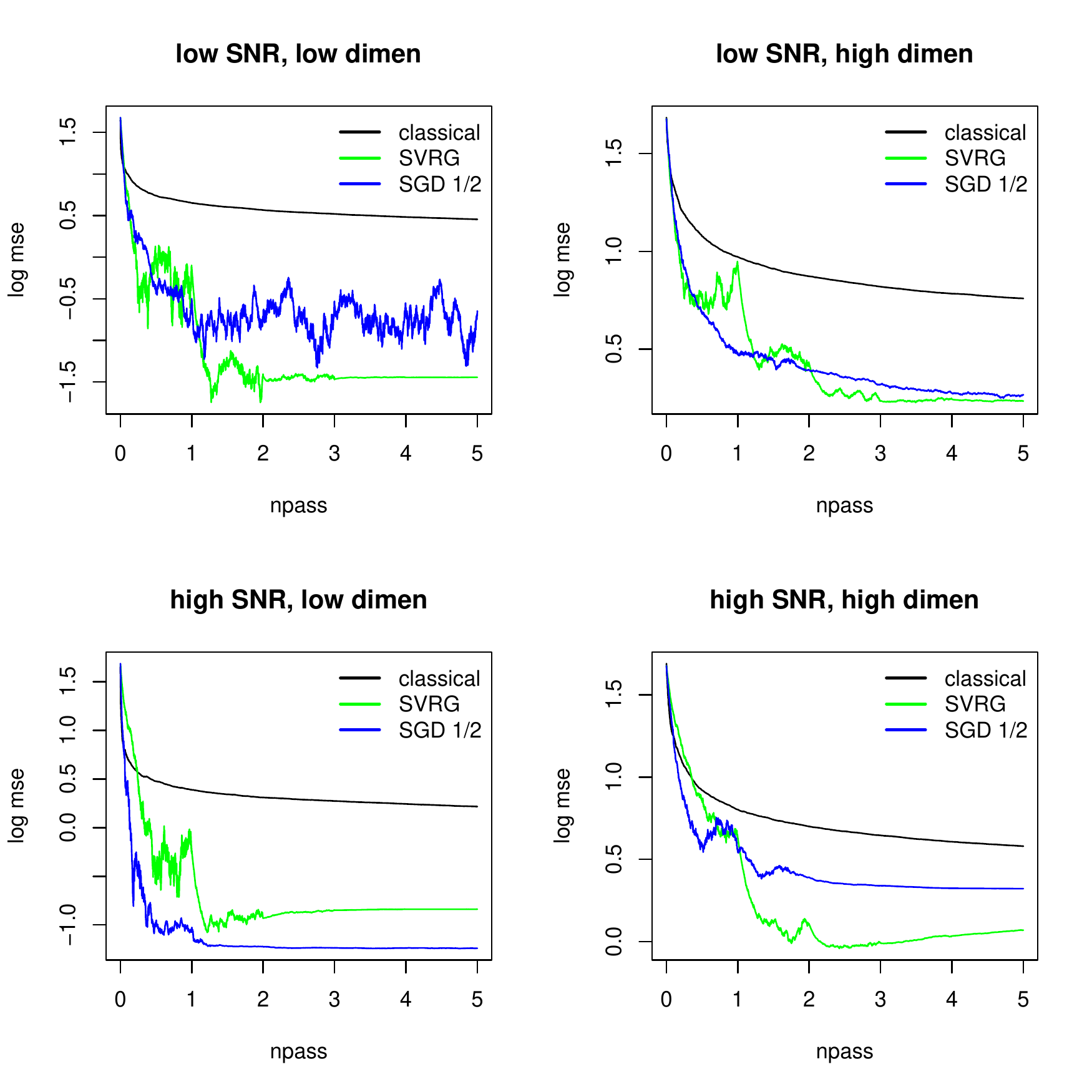} 
\caption{Simulated data experiments, comparing the performance of our procedure \ISGDhalf\ against SVRG and classical ISGD. The left four plots are with the normal model, the right four plots with the logistic model.}
\label{fig:ISGDhalf_synth}
\end{figure*}

From simulations with the normal model in the left half of Figure~\ref{fig:ISGDhalf_synth} we see that \ISGDhalf\ attains a comparable performance to SVRG. In general, SVRG attains an overall better performance for these experiments, which we believe is related to our convergence diagnostic being aggressive in a couple of cases, which are essentially cases of Type-I error.

\begin{figure*}[h!]
\centering
\includegraphics[width=0.49\textwidth]{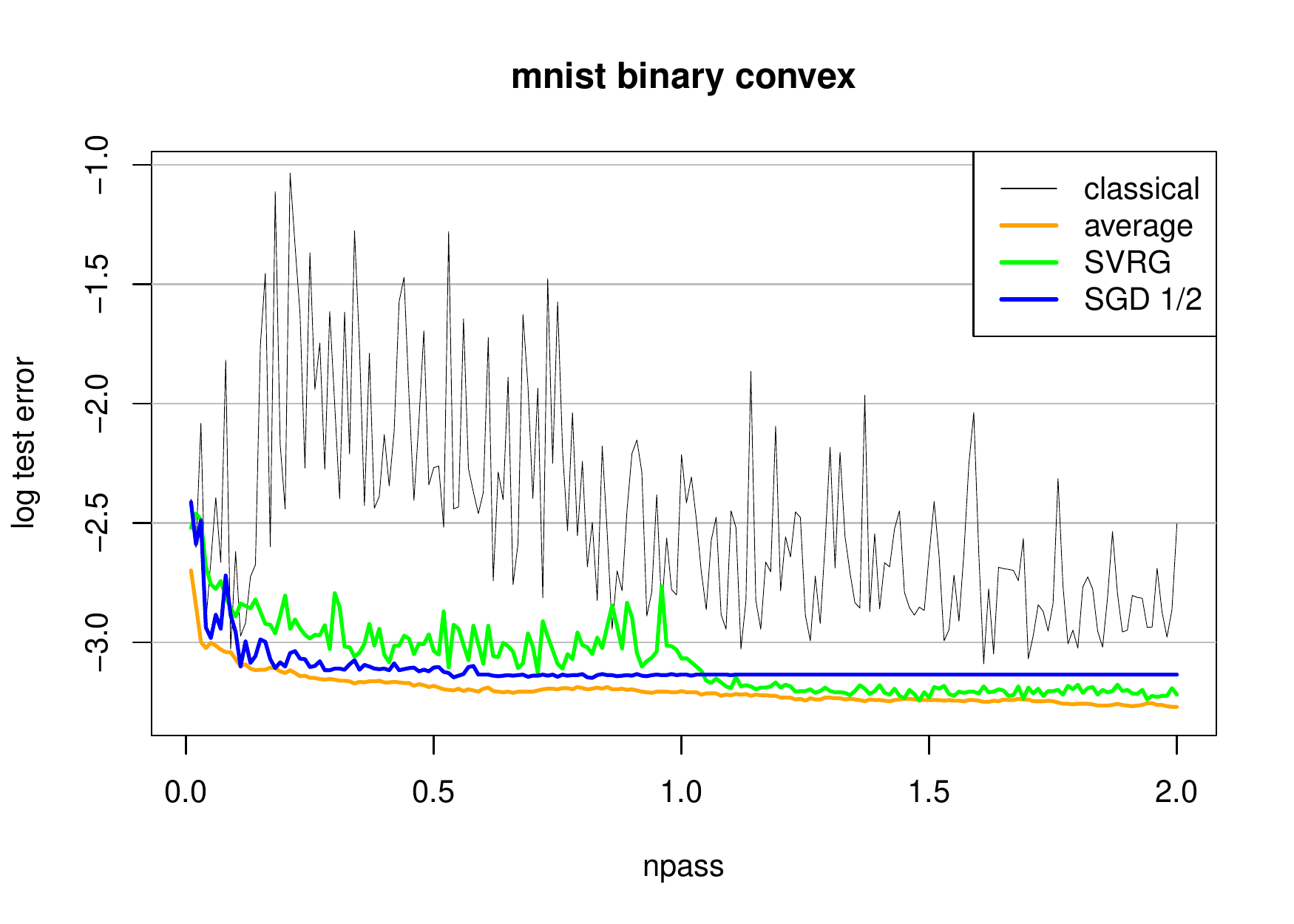}
\includegraphics[width=0.5\textwidth]{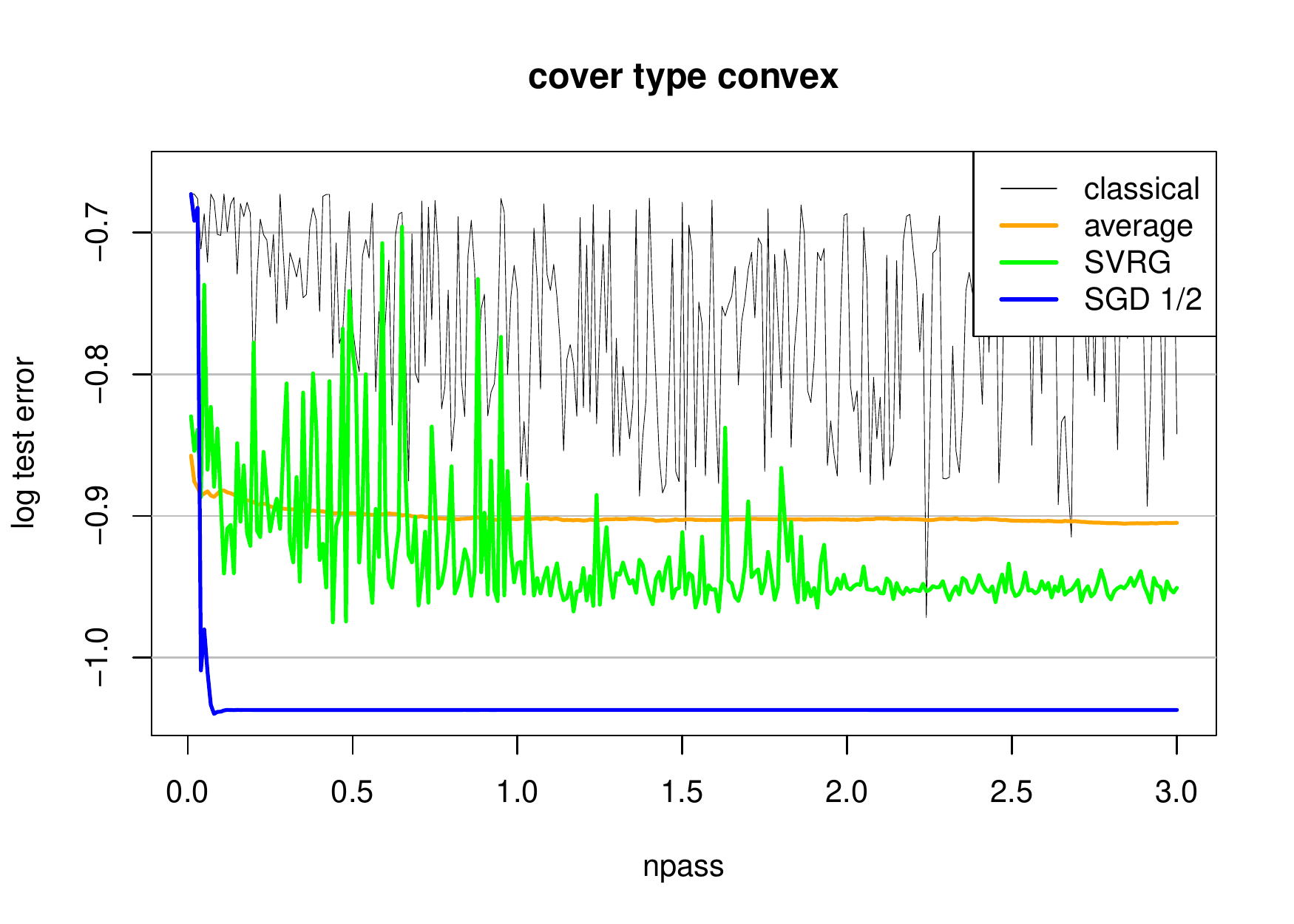} 
\caption{Benchmark data sets with binary logistic regression using \ISGDhalf, SVRG, classical ISGD, and averaged ISGD. Prediction error on a held out test set. MNIST (binary) on the left, COVERTYPE (binary) on the right. }
\label{fig:ISGDhalf_real}
\end{figure*}

From simulations with the logistic model in the right half of Figure~\ref{fig:ISGDhalf_synth} we see that \ISGDhalf \ attains an even better performance than before as there are fewer cases of Type-I error. 
With high SNR and low dimension parameter settings, \ISGDhalf\ achieves consistently better performance than SVRG. We note that such comparisons do not take into account the sensitivity of SVRG to 
misspecifications of the learning rate (large enough learning rates can easily make the procedure diverge); or that SVRG requires periodic calculations over the entire data set, which here is easy because we are using only 5,000 data points, but may be a problem in more realistic settings.
We also note that there are several improvements available for \ISGDhalf\ by allowing a larger {\tt burnin} period or by discounting the learning rate less aggressively.
An interesting direction for future work is to 
understand the performance of our diagnostic test in terms of statistical validity and power, 
and thus address some of the aforementioned tuning issues in a principled manner.

\subsection{Benchmark data sets}
\label{section:ext_experiment_bench}
In addition to simulated experiments we conduct experiments on benchmark data sets  MNIST (binary) and COVERTYPE (binary) to evaluate real world performance.\footnote{Data sets can be found at \url{https://archive.ics.uci.edu/ml/databases/mnist/} and \url{https://archive.ics.uci.edu/ml/datasets/covertype}, respectively.}~In particular, we perform binary logistic regression using \ISGDhalf, SVRG, classical ISGD, and averaged ISGD~\citep{toulis2016towards}. We plot the prediction error on a held-out test set in Figure~\ref{fig:ISGDhalf_real} relative to the number of passes over the data.

Overall, we see that \ISGDhalf\ convergences very quickly, after going over less than a quarter of the data, 
and achieves best performance in the COVERTYPE data set.
We currently do not have a theoretical justification for this, but we have verified that the aforementioned result is consistently observed across multiple experiments.
\ISGDhalf\ was also very stable to specifications of the learning rate parameter, as expected from the analysis of Theorem~\ref{theorem:linear_implicit}.
In contrast, even though SVRG performed comparably to \ISGDhalf, its performance was unstable, especially in the COVERTYPE data set, and required careful fine tuning of the learning rate through trial and error.~Averaged SGD performed well on the MNIST data set, but flattened out very fast in the COVERTYPE data, possibly due to non-strong convexity of the objective function. 


\section{Conclusion}
\label{section:conclusion}
In this paper we focused on detecting convergence of SGD with constant learning rate to its
convergence phase. This is an important practical task 
because statistical properties of iterative stochastic procedures are better understood under stationarity. We borrowed from the theory of stopping times in stochastic approximation to develop a simple diagnostic that uses the inner product of successive gradients to detect convergence. Theoretical and empirical results suggest that the diagnostic reliably detects the phase transition, which can speed up classical procedures.

Future work needs to focus on analysis of errors $||\theta_n-\theta_\star||^2$ conditional on the diagnostic being activated. This could show that the error is uncorrelated with the initial starting point conditional on the test being activated, and so provide theoretical support to the empirical results in Table~\ref{table:diagnostic}. 
%
It would also be interesting to analyzse \ISGDhalf. Another idea is to use aggregation among parallel \ISGDhalf\ chains. At stationarity we expect iterates from different chains to be uncorrelated with each other, 
and so averaging may help. It would also be interesting 
to use the diagnostic in problems with non-convex loss, such as neural networks.

\pagebreak
\singlespacing
\bibliographystyle{chicago}
\bibliography{chee18-arxiv}

\clearpage
\appendix

\doublespacing

\section{Proofs of theorems}
\setcounter{theorem}{0}
\setcounter{proof}{1}

\SGDTheorem

\TheoremPflugFinite
\ProofPflugFinite

{\em Remarks.} Assumption~(b) is a form of strong convexity.
For example, suppose that $y = x^\top\theta_\star+ e$, then 
$f(x, \theta) = x x^\top (\theta-\theta_\star)$ and 
$f(x, \theta-\gamma z)^\top z = f(x, \theta)^\top z
- \gamma z^\top \Ex{x x^\top} z$. In this case 
$C = \Ex{x x^\top}$ is the Fisher information matrix and 
Assumption~(b) holds for $K=1$. When $\gamma$ 
is small enough and a Taylor approximation of $f(x, \theta-\gamma z)$ is possible, the above result still holds for $K=1$ when the Fisher information exists.
Assumption~(c) shows that there is a threshold value for $\gamma$ below which the diagnostic cannot terminate.
For example, suppose that error noise is small so that $\tau^2\approx0$ and $K=1$, as argued before. Then, $\gamma > 1/\mu$, that is, the learning rate has to exceed the reciprocal of the minimum eigenvalue of the Fisher information matrix.

\setcounter{theorem}{2}
\LinearTheorem
\setcounter{proof}{2}
\LinearProof

\LinearTheoremImplicit
\LinearProofImplicit

\TheoremGeneral
\ProofGeneral

\section{Error analysis for constant learning rate ISGD}

In this section, $\ell$ will denote likelihood, which is the negated loss (cf. Equation~\eqref{eq:implicit}).
Thus, we have the implicit update of SGD (ISGD):
\begin{align}
\label{eq:implicit}
\theta_n  = \theta_{n-1} + \gamma \nabla\ell(y_n, x_n^\top\theta_n).
\end{align}

We will operate under the following assumptions:
\Assumptions

To prove Theorem~\ref{theorem:isgd}, our result for the upper bound on the MSE for constant learning rate ISGD, we first prove the following results:
\setcounter{lemma}{5}
\TheoremGradient
\ProofGradient
\LemmaLambda
\ProofLambda

\setcounter{theorem}{7}
\TheoremBound
\ProofBound
\setcounter{lemma}{8}
\LemmaViableLR
\ProofViableLR

\end{document}